\documentclass[a4paper]{article}

\usepackage{INTERSPEECH2016}

\usepackage{graphicx}
\usepackage{amssymb,amsmath,bm}
\usepackage{textcomp}
\usepackage{url}
\usepackage{balance}

\def\vec#1{\ensuremath{\bm{{#1}}}}

\ninept

\title{Calibration of Phone Likelihoods in Automatic Speech Recognition}

\makeatletter
\def\name#1{\gdef\@name{#1\\}}
\makeatother 

\name{{\em David A. van Leeuwen$^{1,2}$ and Joost van Doremalen$^1$}}

\address{$^1$NovoLanguage, Nijmegen, The Netherlands\\
  $^2$CLS/CLST, Radboud University Nijmegen, The Netherlands\\
  {\small \tt david@novolanguage.com, joost@novolanguage.com}
}

\let\underscore=\_
\def\_{\checkmath_\underscore}
\def\checkmath#1#2{\ifmmode\def\next##1{#1{\rm##1}}\else\let\next=#2\fi\next}
\let\circonflex=\^
\def\^{\checkmath^\circonflex}

\def\x{{\vec x}}
\def\wi{\{w_i\}}
\def\xt{\{\x_t\}}
\def\pif{\pi_{\!f}}
\def\Nf{N_{\!f}}
\def\T{{\cal T}}
\def\cllr{\ensuremath{C\_{llr}}}
\def\vlambda{{\vec \lambda}}
\def\vbeta{{\vec \beta}}
\def\mce{\ensuremath{H\_{mc}}}
\def\mcemin{\ensuremath{H\_{mc}\^{min}}}
\def\mcecal{\ensuremath{H\_{mc}\^{cal}}}

\def\fig#1{Fig.~\ref{fig:#1}}

\def\tab#1{Table~\ref{tab:#1}}
\def\eq#1{(\ref{eq:#1})}

\begin{document}

  \maketitle
  \begin{abstract}
    In this paper we study the probabilistic properties of the posteriors in a speech recognition system that uses a deep neural network (DNN) for acoustic modeling.  We do this by reducing Kaldi's DNN shared pdf-id posteriors to phone likelihoods, and using test set forced alignments to evaluate these using a calibration sensitive metric.  Individual frame posteriors are in principle well-calibrated, because the DNN is trained using cross entropy as the objective function, which is a proper scoring rule.  When entire phones are assessed, we observe that it is best to average the log likelihoods over the duration of the phone.  Further scaling of the average log likelihoods by the logarithm of the duration slightly improves the calibration, and this improvement is retained when tested on independent test data.  
  \end{abstract}

  
\section{Introduction}
\label{Introduction}

Automatic Speech Recognition has benefitted from a probabilistic approach for many decades.  Modern architectures based on Deep Neural Networks~\cite{Hinton:2012} still describe the inner workings in a probabilistic framework, combining acoustic model likelihoods from observations with language model probabilities that function as a prior.  For the task of speech recognition \emph{per se} these probabilities are not directly used, but rather the model sequence that produces the highest posterior probability are the direct link with the recognition result.  Formally, the word sequence $\wi$ is chosen that maximizes the posterior probability given the acoustic observations $\xt$:
\begin{equation}
  \label{eq:1}
  \wi = \arg \max_{\wi} \frac{P(\xt\mid \wi) \, P(\wi)}{P(\xt)}.
\end{equation}
Because the prime interest is in the word sequence $\wi$, little attentions is given to the normalizing factor $P(\xt)$, which is not dependent on the word sequence and can hence be ignored in finding the maximum.  The actual probabilistic interpretation of the acoustic and language model then become less relevant, and is primarily retained in the choice of a language model scaling factor~$A$, which weights the relative contribution of the latter to the former.  The scaling of the acoustic likelihood also plays a role in sequence-discriminative training criterions~\cite{Vesely:2013} such as Maximum Mutual Information and Minimum Bayes Risk. 

In the literature, various explanations are given why this scaling factor needs to be there, apart from a simple engineering reason to optimize performance.  In our opinion, the most appealing one~\cite{Gillick:2011} is that in computing acoustic likelihoods, the assumption of frame independence is made, which is obviously incorrect.  Apart from the fact that more often than not frames overlap for over 50\,\%, consecutive frames within the stable portion of a phone will be highly correlated.  Frame independence allows the total log likelihood to be computed as the sum of individual frame log likelihoods.  Taking interframe correlation properly into account would be very hard, so as a remedy the acoustic log likelihood is scaled down by a factor~$1/A$.  But there are other explanations, too. The SPRAAK toolkit's documentation\footnote{\url{http://www.spraak.org/documentation/doxygen/doc/html/index.html}} states that (for HMMs) the acoustic feature space dimension is too way too high, typically 39 where perhaps 10 dimensions would better describe the manyfold in which the speech features are embedded, and this would lead to a power of four overestimation of the acoustic likelihoods.  In \cite{Varona:2004}, the authors give as main reason that acoustic and language model are estimated from different knowledge sources, and therefore need to be combined with their own scaling factor.

In this paper we study the probabilistic properties of phone likelihoods by themselves, i.e., not in relation to language model properties.  Specifically, we investigate how individual frame likelihoods can optimally be combined to phone likelihoods.  We evaluate the quality of the phone likelihoods with the multiclass cross entropy (\mce).  This error metric is \emph{calibration sensitive}, i.e, it penalizes under- or overconfident probabilistic statements.  In speech, a variant of this metric was introduced as $\cllr$ in language recognition~\cite{Brummer:2006a}, where, in NIST evaluation context, the classification problem is cast in a detection framework in order to assess calibration.  

This paper is organized als follows.  First, we discuss a metric for calibration and how we obtain phone likelihoods from DNN posteriors.  Then we present experimental results, before we conclude. 

\section{Acoustic likelihoods}
\label{sec:eval-acoust-likel}

\subsection{Evaluation metric}

Let $\vlambda$ be a vector of phone log likelihoods that are produced by a recognition system for a speech segment $\xt$ over the duration of a phone.  With $N$ phone classes (including non-speech classes), the posterior probability for a specific phone $f$ is 
\begin{equation}
  \label{eq:posterior}
  p(f\mid \xt) = \frac{\pif \, e^{\lambda_f}}{\sum_i \pi_i \, e^{\lambda_i}}, 
\end{equation}
where $\pi_i$ are the priors of the phones.  Please note~\cite{Brummer:2006a}, that $\exp \vlambda$ can be scaled by an arbitrary (positive) factor, and that the log likelihoods span only an $N-1$ dimensional space, because the posteriors and priors both sum up to~1. 

For a collection of labeled phones $\T=\{f_k\}$ in a test set, the cross entropy is defined as
\begin{equation}
  \label{eq:MCE}
  \mce = \frac1N \sum_{f=1}^N \frac1{\Nf} \sum_{k\in \T_f} -\log p(k\mid \xt_k),
\end{equation}
i.e., an average log penalty over the posterior of the true phone class, where the amounts of phones for each class~$\Nf$ are equalized.  This penalty is a \emph{proper scoring rule}, and has the property that more certainty towards the true class reduces the penalty, but that expressing similar certainty towards the wrong class increases the penalty by a much larger amount.  It is therefore important that the likelihoods are well scaled w.r.t.\ each other. 

\subsection{Calibration}

\mce\ is calibration sensitive, but how can we determine what part of \mce\ is due to calibration errors and what part due to discrimination errors?  For two-class systems, such as in speaker recognition with `target' and `non-target' speaker classes, this separation can be determined exactly~\cite{Brummer:2006}\footnote{In \cite{Brummer:2006} a scaled variant of \mce\ is named \cllr, the cost of the log-likelihood-ratio.}.  The space spanned by $\vlambda$ is then only one-dimensional, and it is common to use the log likelihood ratio as the single speaker-comparison score.  For a set of supervised trials, the optimal score-to-log-likelihood-ratio mapping can be determined using isotonic regression, and the \mce\ computed after this mapping can be considered optimal in terms of calibration, given the discrimination ability of the recognizer.  The difference between the real \mce\ and the optimal \mce\ is called the calibration cost. 

The generalization of this `2-class minimum \mce' to the multi-class situation is not straightforward~\cite{Brummer-PhD:2010}, but in \cite{Brummer:2006a} it is suggested that we can apply an affine transform the log likelihood vector, i.e.
\begin{equation}
  \label{eq:calibration}
  \vlambda' = \alpha \vlambda + \vbeta, 
\end{equation}
where $\alpha$ is a scaling factor and $\vbeta$ is a vector of offsets.  By optimizing $\alpha$ and $\vbeta$ for minimal \mcemin\ on the test data, we have an upper bound for an `oracle' minimum \mce.  Please note, that the affine transform does not change the discrimination performance between any two phones, so it can be considered a transform that only influences calibration. 

\subsection{From posteriors to log likelihoods}
\label{sec:likel-from-post}

With generative acoustic models, such as HMMs with GMM output probability density functions, the vector $\vlambda$ can be obtained directly from the models for the speech frames under consideration.  For discriminative models, such as DNNs, the likelihoods need to be derived from the posteriors that are formed by the output layer of the DNN~\cite{Robinson:1994}.  In this paper we use a Kaldi~\cite{Povey:2011} standard recipe for acoustic modelling.  In this case, the targets of the DNN are `pdf-id' posteriors, where pdf-ids are shared Gaussians of HMM states.  These states are found by a data driven decision tree training, in such a way that various conditional forms of a phone model (conditioned by phone context, phone position in word, and possibly stress marking or tone) can share these pdf-ids, but no pds-ids are shared between different base phones.  

In order to arrive at phone likelihoods, we will first sum the pdf-id posteriors $p(i \mid \x)$ and priors $p(i)$, where $i$ denotes a pdf-id, over all pdf-ids that contribute to the same base phone, independent of HMM state, phone context, etc:
\begin{equation}
  \label{eq:phone-post}
  p(f \mid \x) = \sum_{i\in f} p(i\mid \x); \qquad p(f) = \sum_{i\in f} p(i),
\end{equation}
where we loosely write $i \in f$ to indicate the mapping of pdf-id $i$ to base phone~$f$.   Please note, that the phone priors $p(f)$ are the priors obtained from the acoustic model, roughly the proportions of frames in training associated with the respective phones, which can be different from the priors~$\pif$ used in \eq{posterior}--\eq{MCE}.  

Once we have the phone posteriors and priors, we can compute the log likelihood for a speech frame $\x$ simply as 
\begin{equation}
  \label{eq:phone-llh}
  \lambda_f(\x) = \log \frac{p(f \mid \x)}{p(i)} + \rm const.
\end{equation}

\subsection{From frame- to phone-likelihoods}

The objective function during DNN training in Kaldi is frame-based cross entropy~\cite{Vesely:2013}.  The posteriors, and therefore the derived log likelihoods, are therefore naturally well calibrated at the frame level.  However, we are interested in likelihoods at the phone level, so the question arises how to go from frame likelihoods to likelihoods for the acoustic observations spanning the entire phone.  In this paper, we assume that the phone segmentation of the test data is given.  There are many ways of combining frame-likelihoods into phone-likelihoods, we will study three:
\begin{enumerate}
\item Use the sum of the frame log likelihoods.  Typically, in ASR decoding, the sum of the log likelihoods is used, suggesting frame independence (i.e., likelihoods can be multiplied)
\item Use the mean of the frame log likelihoods.  This suggests that the frame likelihoods are fully correlated within one phone, and the main effect of averaging is to reduce the noise of the likelihoods. 
\item Use the mean of the frame log likelihoods, multiplied by log-duration.  This is a middle ground between the first two ways of combining frame likelihoods.  The factor \emph{log duration} conservatively appreciates the fact that phones of longer duration have more acoustic evidence and should therefore induce a larger variation of log likelihoods over the different phones.  
\end{enumerate}

\section{Experiments}

\subsection{ASR system and data}
\label{sec:asr-system-data}

For this paper, we use LDC's Wall Street Journal data bases WSJ0 and WSJ1 for training acoustic models, calibration and testing.  We use the standard segmentation of train- and test speakers, which contain 8 speakers in the `eval92' portion from WSJ0 and 10 speakers in the `dev93' portion from WSJ1.  Acoustic models are trained using the \texttt{wsj/s5} recipe of Kaldi~\cite{Povey:2011}.  Specifically, we use the `online nnet2' DNN training, and use the simulated online decoding of the test data, resetting the (i-vector) speaker adaptation information at every utterance.  

We obtain the forced aligned phone segmentations using a cross-word triphone Viterbi decoder, using the canonical pronunciations from the CMU American English dictionary\footnote{A. Rudnicky, \url{http://www.speech.cs.cmu.edu/cgi-bin/cmudict}}. We then compute the phone likelihoods from the DNN outputs using \eq{phone-post} and \eq{phone-llh}, using the trained pdf-id priors.  We then determine the \mce\ over the test set using \eq{posterior} and \eq{MCE}. Because we are purely interested in the discrimination and calibration of the acoustic part of the model, we use a flat prior $\pif=1/N$ in computing \mce.  

\subsection{Multiclass cross entropy}
\label{sec:experiments}

\begin{table}
  \centering
  \begin{tabular}{|l|c|c|c|}
    \hline
    Function& \mce& \mcemin&$\alpha$\\
    \hline
    \multicolumn4{c}{Eval '92}\\
    \hline
    $\sum_t \lambda_t$&            1.081& 0.260& 0.162\\
    $\sum_t \lambda_t / n$&        0.261& 0.232& 1.073\\
    $\log n \sum_t \lambda_t / n$& 0.309& 0.226& 0.586\\
    \hline
    \multicolumn4{c}{Dev '93}\\
    \hline
    $\sum_t \lambda_t$&            1.411& 0.344& 0.160\\
    $\sum_t \lambda_t / n$&        0.350& 0.314& 1.017\\
    $\log n \sum_t \lambda_t / n$& 0.418& 0.305& 0.567\\
    \hline
  \end{tabular}
  \caption{\protect\mce\ results for the `eval92' and `dev93' dataset of WSJ, for the three functions that combine frame likelihoods $\lambda_t$ into a phone likelihood.  The index $t$ runs over the $n$ frames belonging to the phone.  The column $\alpha$ indicates the scaling used in \eq{calibration}.}
  \label{tab:res1}
\end{table}

In \tab{res1} we have tabulated the results of measuring \mce\ and \mcemin\ for the two WSJ test sets `eval92' and `dev93.'  We can observe that the log likelihood computed as the sum of frame log likelihoods is not well calibrated naturally, as \mcemin\ is much smaller than \mce---this is the calibration loss.  Also, the scaling factor, needed in the affine transformation of log likelihoods  in order to obtain \mcemin, is 0.16.  This value (approximately 1/6) is not too far off from the acoustic likelihood scaling factors found in large vocabulary ASR.  

When the phone log likelihood is computed as the mean, the natural calibration is much better---it can only slightly be improved by optimizing the parameters of \eq{calibration}.  Also, the scaling factor after optimizing is close to one, which corroborates the natural calibration.  

Finally, when the phone log likelihood is computed as the mean over frames scaled with the log of the number of frames, there is less `natural calibration' (a bigger difference between \mce\ and \mcemin, $\alpha$ not close to 1).  But there is more potential in obtaining a lower \mcemin\ than for the simpler functions.  

\begin{table}[t]
  \centering
  \begin{tabular}{|l|c|c|}
    \hline
    Function& \mce& \mcecal\\
    \hline
    \multicolumn3{c}{Eval '92}\\
    \hline
    $\sum_t \lambda_t$&            1.081& 0.277\\
    $\sum_t \lambda_t / n$&        0.261& 0.247\\
    $\log n \sum_t \lambda_t / n$& 0.309& 0.242\\                                   
    \hline
    \multicolumn3{c}{Dev '93}\\
    \hline
    $\sum_t \lambda_t$&            1.411& 0.361\\
    $\sum_t \lambda_t / n$&        0.350& 0.332\\
    $\log n \sum_t \lambda_t / n$& 0.418& 0.324\\
    \hline
  \end{tabular}
  \caption{\protect\mce\ values without calibration (\protect\mce) and with calibration (\protect\mcecal), where the calibration parameters are obtained from the other test set.}
  \label{tab:res2}
\end{table}

\subsection{Calibration}
\label{sec:calibration}

The column \mcemin\ in \tab{res1} is obtained by `self calibration,' i.e., using the test data itself to find the optimum calibration parameters $\alpha$ and $\vbeta$.  It is computed only to give an indication of the calibration loss.  But we can also take the optimum parameters found for one test set, and use these to calibrate the log likelihoods of the other test set.  We have indicated the results of this experiment in \tab{res2}.  One can observe from the values in column \mcecal, that the calibration parameters from the other test set improve \mce.  The value is not quite as low as the self-calibration values from \tab{res1}, but this is expected when changing from one data set to the next. 

\begin{figure*}
  \centering
  \includegraphics[width=\hsize]{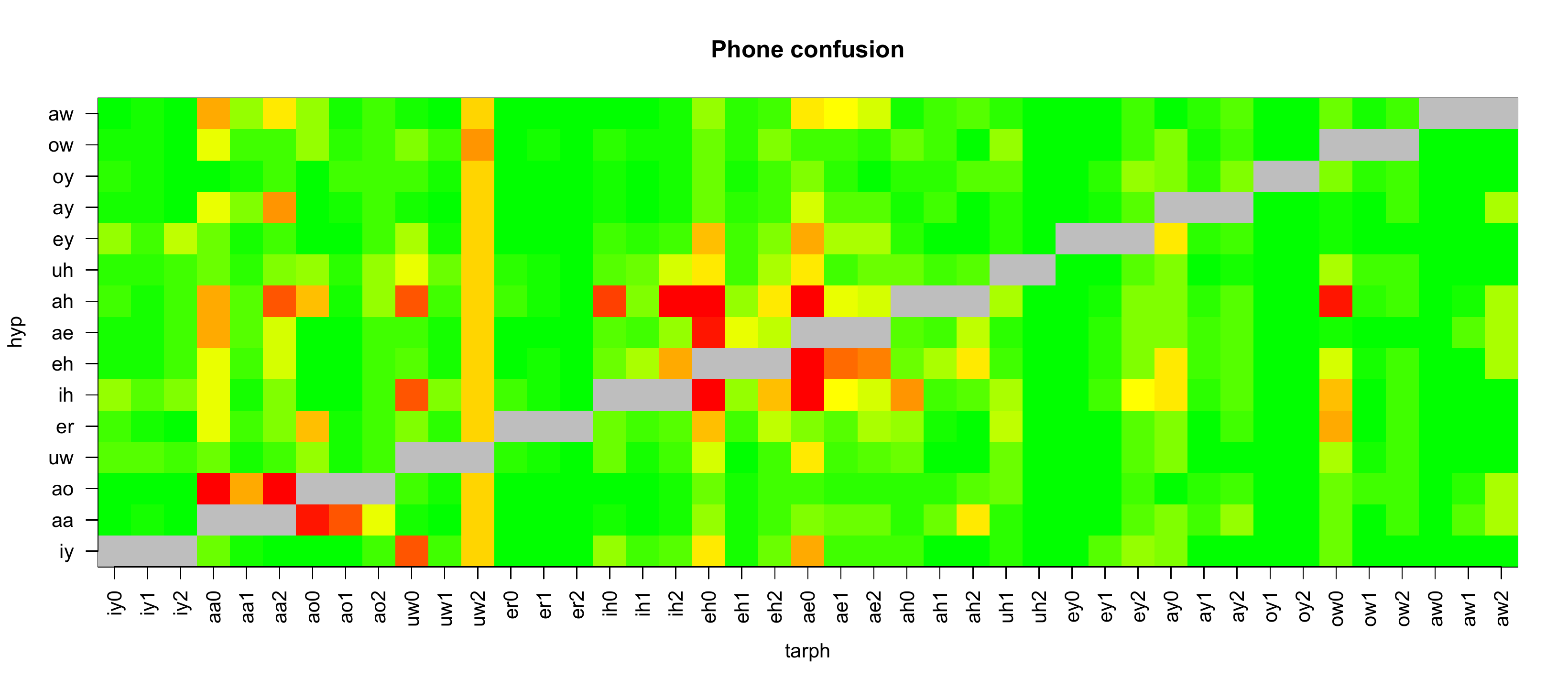}
  \caption{A confusion matrix for the vowels.  Green indicates an EER close to zero, red represents EERs of 25\,\% and above.  Vertical correlation of colour is caused by low target likelihoods for vowels with only a few examples in the test data.}
  \label{fig:vowels}
\end{figure*}

\subsection{A caveat}
\label{sec:caveat}

In these experiments we use a recognizer to produce a forced alignment, and then use this alignment as truth labels for assessing the same recognizer's acoustic models.  We can imagine that if the posteriors vectors show very little uncertainty, i.e., are close to one for just one phone, and vanish for the others, the Viterbi decoding will find a path with low \mce\ regardless of a correct alignment.  In order to test if our measurement of \mce\ is not simply a self-fulfilling prophecy, we force-aligned incorrect transcriptions from `half the database away' to the audio, and computed \mce\ as before.  This gave $\mce = 38$, and after self-calibration a value of 3.3.  This value is just below $\mce = -\log(1/42) \approx 3.7$, the reference value of a flat posterior for a phoneset with $N = 42$ phones.

\subsection{Phone confusion matrix}

\begin{figure}[ht]
  \centering
  \includegraphics[width=\hsize]{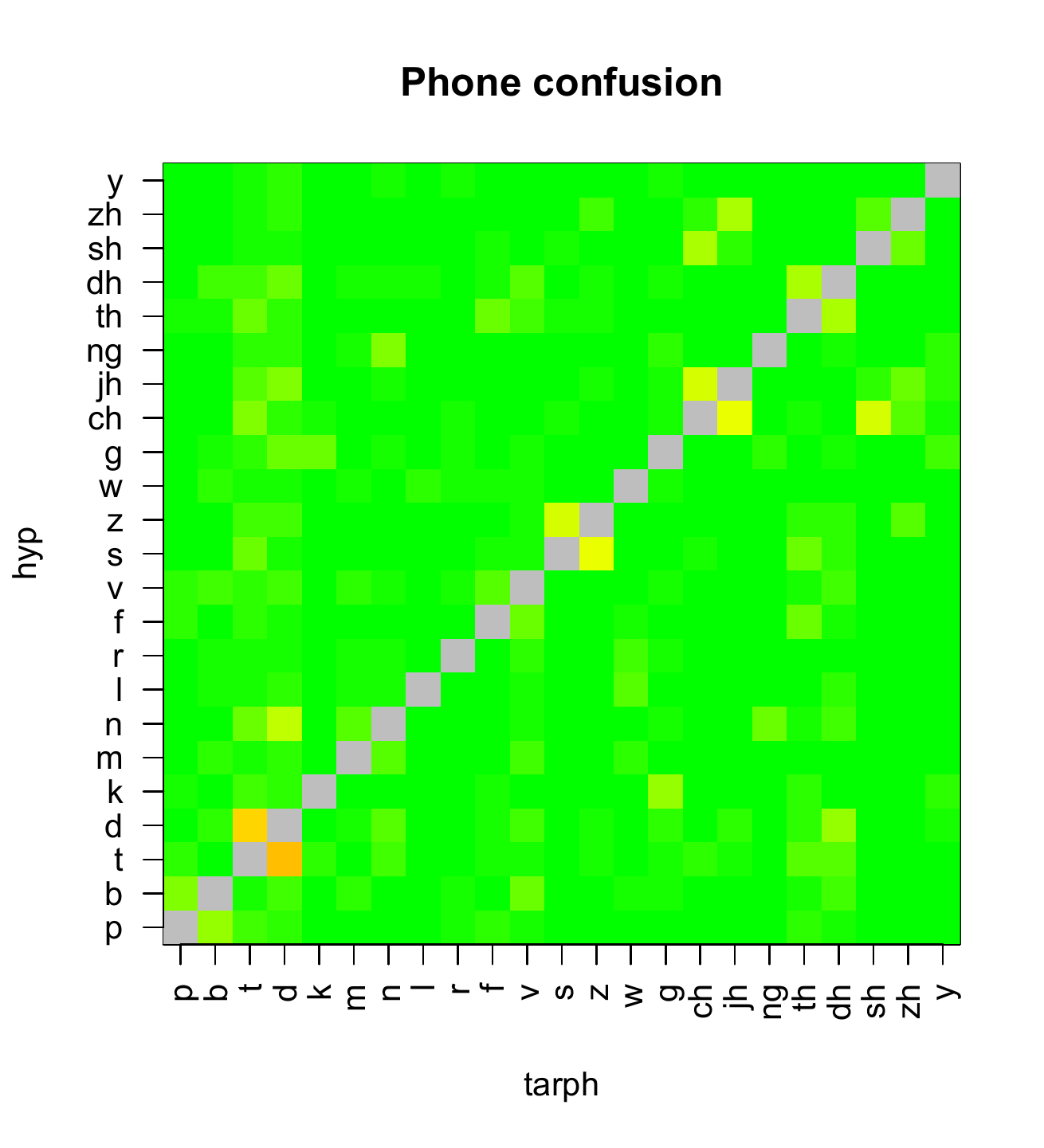}
  \caption{A confusion matrix for the consonants.  The colours are in the same scale as \protect\fig{vowels}}
  \label{fig:consonants}
\end{figure}

The experiments in this section allow for another diagnostic analysis of the acoustic model, namely a form of confusion matrix.  Instead of considering all alternative phones simultaneously, we can just look at a single alternative phone, and determine the discrimination capability of the model between target and alternative phone.  A good discrimination measure is the Equal Error Rate (EER), and the results for the vowels in the `dev93' test set (the harder of the two sets) is shown in \fig{vowels}.  In the CMU dictionary, we have stress indicators for the vowels, so we can analyse the discrimination ability for target vowels with different stress.  In Kaldi, the pdf-ids are shared between different stress conditions of the vowels, so we can't distinguish stress for hypothesis phones.  One thing the figure shows is that unstressed vowels are harder to discriminate than stressed vowels.  Although this paper is mainly about calibration, we show the phone confusion matrix for consonants in \fig{consonants}.

\section{Conclusions}

We have studied the probabilistic properties of a DNN acoustic model.  Because DNNs are trained with a cross entropy objective function, it is not surprising that at the phone level the posteriors are well calibrated.  It seems that averaging the obtained frame log likelihoods over the entire duration of the phone gives a good calibration of the log likelihoods at the phone level.  The average duration of the phones are between 5.9 (for /ah/) and 16.2 (for /aw/) frames of 10\,ms, this is in the same ballpark as the acoustic model scale that is employed in LVCSR.  The main difference is that in LVCSR the acoustic scale factor is applied to each frame equally, whereas in our analysis the scale factor is applied to each phone equally.  The application of calibrated phone likelihoods is perhaps not so much in LVCSR decoding, where the phone alignement itself is part of the task, but more in detailed analysis of pronunciation variation when the utterance text is known a priori.  

We have seen that calibrating the phone likelihoods on one data set does carry over to the next, but of course the two data sets used here are very similar.  It remains to be seen what is left of such calibration if we switch to a different domain.  The transform~\eq{calibration} contains quite a lot of parameters, it is likely that we need to regularize these to make such a calibration more robust.  Perhaps best is to find likelihood combination functions that have a natural tendency to be well-calibrated.  The mean over frame log likelihoods is a first candidate for this---it can hardly be improved by further calibration.  Although the combination function that scales the log likelihood with log duration has a slightly better potential for \mcemin, it is naturally less well calibrated.

\parskip=0pt plus 2pt

\bibliographystyle{IEEEtran}
\bibliography{david-bibdesk}

\end{document}